\documentclass[twoside,twocolumn,english]{paper}
\usepackage[T1]{fontenc}
\usepackage[latin9]{inputenc}
\usepackage[a4paper]{geometry}
\usepackage{babel}
\usepackage{array}
\usepackage{textcomp}
\usepackage{url}
\usepackage{multirow}
\usepackage{graphicx}
\usepackage[unicode=true]
 {hyperref}

\makeatletter

\providecommand{\tabularnewline}{\\}

\usepackage{float}
\floatstyle{ruled} 
\restylefloat{figure}

\makeatother

\begin{document}

\title{3D Terrestrial lidar data classification of complex natural scenes
using a multi-scale dimensionality criterion: applications in geomorphology.}

\author{Nicolas Brodu$^{a}$, Dimitri Lague$^{a,b}$}

\institution{a: Geosciences Rennes, Université Rennes 1, CNRS, Rennes, France.\\
b: Dpt of Geological Sciences, University of Canterbury, Christchurch,
New-Zealand.}
\maketitle
\begin{abstract}
3D point clouds of natural environments relevant to problems in geomorphology
(rivers, coastal environments, cliffs,...) often require classification
of the data into elementary relevant classes. A typical example is
the separation of riparian vegetation from ground in fluvial environments,
the distinction between fresh surfaces and rockfall in cliff environments,
or more generally the classification of surfaces according to their
morphology (e.g. the presence of bedforms or by grain size). Natural
surfaces are heterogeneous and their distinctive properties are seldom
defined at a unique scale, prompting the use of multi-scale criteria
to achieve a high degree of classification success. We have thus defined
a multi-scale measure of the point cloud dimensionality around each
point. The dimensionality characterizes the local 3D organization
of the point cloud within spheres centered on the measured points
and varies from being 1D (points set along a line), 2D (points forming
a plane) to the full 3D volume. By varying the diameter of the sphere,
we can thus monitor how the local cloud geometry behaves across scales.
We present the technique and illustrate its efficiency in separating
riparian vegetation from ground and classifying a mountain stream
as vegetation, rock, gravel or water surface. In these two cases,
separating the vegetation from ground or other classes achieve accuracy
larger than 98 \%. Comparison with a single scale approach shows the
superiority of the multi-scale analysis in enhancing class separability
and spatial resolution of the classification. Scenes between ten and
one hundred million points can be classified on a common laptop in
a reasonable time. The technique is robust to missing data, shadow
zones and changes in point density within the scene. The classification
is fast and accurate and can account for some degree of intra-class
morphological variability such as different vegetation types. A probabilistic
confidence in the classification result is given at each point, allowing
the user to remove the points for which the classification is uncertain.
The process can be both fully automated (minimal user input once,
all scenes treated in large computation batches), but also fully customized
by the user including a graphical definition of the classifiers if
so desired. Working classifiers can be exchanged between users independently
of the instrument used to acquire the data avoiding the need to go
through full training of the classifier. Although developed for fully
3D data, the method can be readily applied to 2.5D airborne lidar
data.
\end{abstract}

\section{Introduction}

Terrestrial laser scanning (TLS) is now frequently used in earth sciences
studies to achieve greater precision and completeness in surveying
natural environments than what was feasible a few years ago. Having
an almost complete and precise documentation of natural surfaces has
opened up several new scientific applications. These include the detailed
analysis of geometric properties of natural surfaces over a wide range
of scales (from a few cm to km), such as 3D stratigraphic reconstruction
and outcrop analysis \cite{Labourdette07,Renard06}, grain size distribution
in rivers \cite{HodgeESPL09,HodgeSedim09,Heritage2009_grainroughness},
dune fields\cite{Nield11,Dune04}, vegetation hydraulic roughness
\cite{Antonarakis10,Antonorakis09}, channel bed dynamics \cite{Milan_proglacialchange2007}
and in situ monitoring of cliff erosion and rockfall characteristics
\cite{Abellan06,Lim09,Rabatel08,Rosser05,Teza08}.For all these applications,
precise automated classification procedures that can pre-process complex
3D point cloud in a variety of natural environments are highly desirable.
Typical examples of applications are the separation of vegetation
from ground or cliff outcrops, the distinction between fresh rock
surfaces and rockfall, the classification of flat or rippled bed and
more generally the classification of surfaces according to their morphology.
Yet, developing such procedures in the context of geomorphologic applications
remains difficult for four reasons : (1) the 3D nature of the data
as opposed to the traditional 2D structures of digital elevation models
(DEM), (2) the variable degree of resolution and completeness of the
data due to inevitable shadowing effects, (3) the natural heterogeneity
and complexity of natural surfaces, and (4) the large amount of data
that is now generated by modern TLS. In the following we describe
these difficulties and how efficient 3D classification is critically
needed to advance our use of TLS data in natural environments.

\begin{figure*}[!t]
\begin{centering}
\includegraphics[width=1\textwidth]{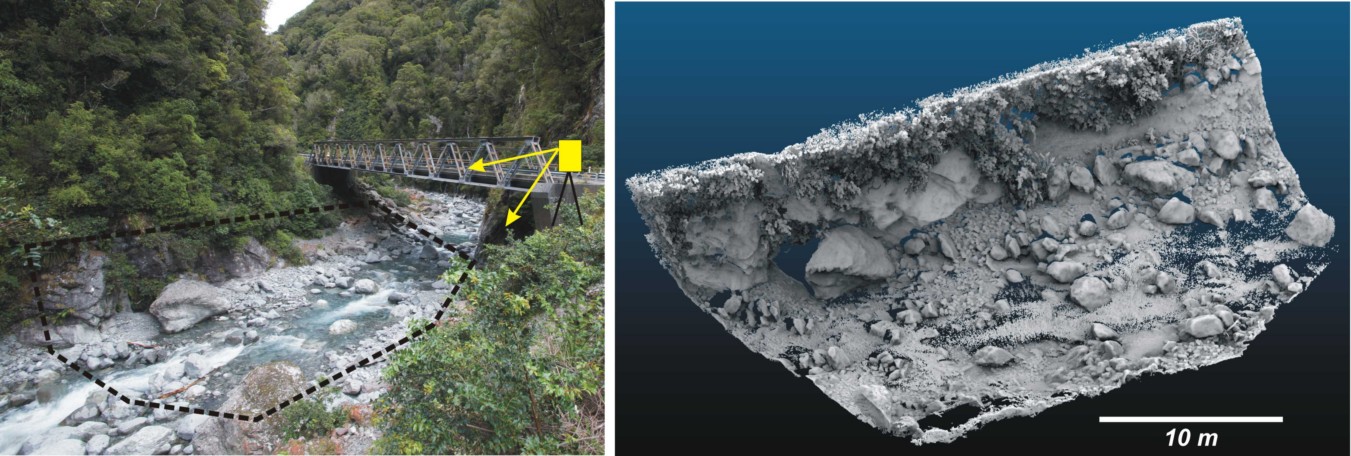}
\par\end{centering}

\caption{Left : Steep mountain river bed in the Otira gorge (New-Zealand),
and Terrestrial Laser Scanner location. Right: part of the point cloud
rendered using PCV technique in CloudCompare \cite{CloudCompare}
showing the full 3D nature of the scene (3 millions points, minimum
point spacing = 1 cm). Identifying key elementary classes such as
vegetation, rock surfaces, gravels or water surfaces would allow to
study the vertical distribution of vegetation, the water surface profile,
to segment large boulders, or to measure changes in gravel cover and
thickness between surveys.\label{fig:Otira-river-bed}}
\end{figure*}

1. Terrestrial lidar data are mostly 3D as opposed to digital elevation
models or airborne lidar data which can be considered 2.5D. This means
that traditional data analysis methods based on raster formats (in
particular the separation of vegetation from ground, e.g. \cite{BareEarth})
or 2D vector data processing cannot in general be applied to ground
based lidar data. In some cases, the studied area in the 3D point
cloud is mostly 2D at the scale of interest (i.e., river bed \cite{HodgeESPL09},
cliff \cite{Rosser05,Abellan10}, estuaries \cite{Guarnieri09}) and
can be projected and gridded to use existing fast raster based methods.
However in many cases the natural surface is 3D and there is no simple
way to turn it into a 2D surface (e.g., Fig. \ref{fig:Otira-river-bed}).
In other cases rasterizing a large scale 2D surface becomes non-trivial
when sub-pixel features (vegetation, gravel, fractures...) are significantly
3D. In a river bed for instance, this amounts at locally classifying
the data in terms of bed surface and over-bed features (typically
vegetation) which requires a 3D classification approach. 

2. Terrestrial lidar datasets are all prone to a variable degree to
shadow effects and missing data (water surface for instance) inherent
to the ground based location of the sensor and the roughness characteristics
of natural surfaces (e.g. \ref{fig:Otira-river-bed}). While multiple
scanning positions can significantly reduce this issue, it is sometimes
not feasible in the field due to limited access or time. Interpolation
can be used to fill in missing information (e.g., meshing the surface),
but it is quite complicated in 3D, and can lead to spurious results
owing to the high geometrical complexity of natural surfaces. Arguably,
interpolation should be used as a last resort, and in particular only
after the 3D scene has been correctly classified to remove, for instance,
vegetation. Hence, any method to classify 3D point clouds should account
for shadow effects, either by being insensitive to it, or by factoring
in that data are locally missing.

3. As shown in a scan of a steep mountain stream, natural surfaces
can exhibit complex geometries (fig. \ref{fig:Otira-river-bed}).
This complexity arises from the non-uniformity of individual objects
(variable grain size, type and age of vegetation, variable lithology
and fracture density ...), the large range of characteristic spatial
scales (from sand to boulders, grass to trees) or its absence (fractures
for instance). This makes the classification of raw 3D point cloud
data arguably more complex than artificial structures such as roads
or buildings which have simpler geometrical characteristics (e.g.,
plane surface or sharp angles)

4. As technology evolves, data sets are denser and larger which means
that projects with billions of points are likely to become common
in the next decade. Automatic processing is thus urgently needed,
together with fast and precise methods minimizing user input for rapidly
classifying large 3D points clouds.

To our knowledge no technique has been proposed to classify natural
3D scenes as complex as the one in fig. \ref{fig:Otira-river-bed}
into elementary categories such as vegetation, rock surface, gravels
and water. Classification of simpler environments into flat surfaces
and vegetation has been studied for ground robot navigation \cite{Vandapel04,Lalonde06}
using purely geometrical methods, but was limited by the difficulty
in choosing a specific spatial scale at which natural geometrical
features must be analyzed. Classification based on the reflected laser
intensity has recently been proposed \cite{Franceschi09}, but owing
to the difficulty in correcting precisely for distance and incidence
effects (e.g. \cite{Correction_methods_intensity_2011,Lichti05}),
it cannot yet be applied to 3D surfaces. Classification based on RGB
imagery can be used in simple configurations to separate vegetation
from ground for instance \cite{Lichti05}. But for large complex 3D
environment, the classification efficiency is limited by strong shadow
projections (fig. \ref{fig:Otira-river-bed}), image exposure variations,
effects of surface humidity as well as the limited separability of
spectral signature of RGB components \cite{Lichti05}. Moreover, when
the objective is to classify objects of similar RGB characteristics
but different geometrical characteristics (i.e. flat bed vs ripples,
fresh bedrock vs rockfall), only geometry can be used to separate
points belonging to each class.

In this paper, we present a new classification method for 3D point
clouds specifically tailored for complex natural environments. It
overcomes most of the difficulties discussed above: it is truly 3D,
works directly on point clouds, is largely insensitive to shadow effects
or changes in point density, and most importantly it allows some degree
of variability and heterogeneity in the class characteristics. The
set of softwares designed for this task (the CANUPO suite) is coded
to handle large point cloud datasets. This tool can be used simply
by non-specialists of machine learning both in an automated way and
also by allowing an easy control of the classification process. Because
geometrical measurements are independent of the instrument used (which
is not the case for reflected intensity \cite{Correction_methods_intensity_2011}or
RGB data), classifiers defined in a given setting (i.e. mountain rivers,
salt marsh environment, gravel bed river, cliff outcrop...) can be
directly reused by other users and with different instruments without
a mandatory phase of classifier reconstruction.

The strength of our method is to propose a reliable classification
of the scene elements based uniquely on their 3D geometrical properties
across multiple scales. This allows for example recognition of the
vegetation on complex scenes with very high accuracy (i.e. $\approx99.6\%$
in a context such as fig. \ref{fig:Otira-river-bed}). We first present
the study sites and data acquisition procedure. We then introduce
the new multi-scale dimensionality feature that is used to describe
the local geometry of a point in the scene and how it can characterizes
simple elementary environment features (ground and vegetation). In
section 4, we describe the training approach to construct a classifier:
it aims at automatically finding the combination of scales that best
allows the distinction between two or more features. The quality of
the classification method is tested on two data sets: a simple case
of riparian vegetation above sand, and a more complex, multiple class
case of a mountain river with very pronounced heterogeneity and 3D
features (fig. \ref{fig:Otira-river-bed}). Finally, we discuss the
limitation and range of application of this method with respect to
other classification methods.

\section{\label{sec:Study-sites-and}Study sites and data acquisition}

The method is tested on two different environments : a pioneer salt
marsh environment in the Bay of Mont Saint-Michel (France) scanned
at low tide consisting of riparian vegetation of 10 to 30 cm high
above a sandy ground either flat or with ripples of a few cm height
(fig. \ref{fig:Density-plots} and \ref{fig:bench_visu}); and a steep
section of the Otira River gorge (South Island of New-Zealand) consisting
of bedrock banks partially covered by vegetation and an alluvial bed
composed of gravels and blocks of centimeter to meter size (fig. \ref{fig:Otira-river-bed}).
Both scenes were scanned using a Leica Scanstation 2 mounted on a
survey tripod at 2 m above ground in the pioneer riparian vegetation
or on the bank as in figure \ref{fig:Otira-river-bed} for the Otira
River. The Leica Scanstation 2 is a single echo time-of-flight lidar
using a green laser (532 nm) with a practical range on natural surfaces
varying from 100 to 200 m depending on surface reflectivity. When
the laser incidence is normal to an immobile water surface, the laser
can penetrate up to 30 cm in clear water and return an echo from the
channel bed. This was the case in some part of the Otira Gorge scene.
However, on turbulent white water, the laser is directly reflected
from the surface or penetrates partially the water column\cite{Milan10}.
Hence, the water surface becomes visible as highly uncorrelated noisy
surface (fig. \ref{fig:Otira-river-bed}). Quoted accuracy from the
constructor given as one standard deviation at 50 m are 4 mm for range
measurement and 60 \textmu{}rad for angular accuracy. Repeatability
of the measurement at 50 m was measured at 1.4 mm on our scanner (given
as one standard deviation). Laser footprint is quoted at 4 mm between
1 and 50 m. This narrow footprint allows the laser to hit ground or
cliff point in relatively sparse vegetation. But this also generates
a small proportion of spurious points called mixed-point (e.g. \cite{HodgeESPL09,Lichti07})
at the edges of objects (gravels, stems, leaves ....). The impact
of these spurious points on the classification procedure is addressed
in the discussion section.

Point clouds used for the tests were acquired from a single scan position
as it corresponds to the worst case scenario with respect to shadow
effects and change in point density. In the Otira River, the horizontal
and vertical angular resolution were (0.031\textdegree{}, 0.019\textdegree{})
with a range of distance from the scanner from 15 to 45 m. This corresponds
to point spacing ranging from 5 to 24 mm. To speed up calculation
during the classification tests, the data were sub-sampled with a
minimum point distance of 10 mm leaving 1.17 million points in the
scene. Parameters for the riparian vegetation environment were (0.05\textdegree{},0.014\textdegree{})
for the angular resolution and a distance of 10 to 15 m from the scanner.
This corresponds to point spacing varying from 2.4 mm to 13 mm for
about 640000 points in the dataset used for classification tests.
No further treatment was applied to the data.

\section{Multi-scale local dimensionality feature}

The main idea behind this feature is to characterize the local dimensionality
properties of the scene at each point and at different scales. By
``local dimensionality'' we mean here how the cloud geometrically
looks like at a given location and a given scale: whether it is more
like a line (1D), a plane surface (2D), or whether points are distributed
in the whole volume around the considered location (3D). For instance,
consider a scene comprising a rock surface, gravels, and vegetation
(e.g. fig. \ref{fig:Otira-river-bed}): at a few centimeter scale
the bedrock looks like a 2D surface, the gravels look 3D, and the
vegetation is a mixture of elements like stems (1D) and leaves (2D).
At a larger scale (i.e. 30 cm) the bedrock still looks mostly 2D,
the gravels now look more 2D than 3D, and the vegetation has become
a 3D bush (see fig \ref{Fig:Diag_ternaire_OTIRA}). When combining
information from different scales we can thus build signatures that
identify some categories of objects in the scene. Within the context
of this classification method, the signatures are defined automatically
during the training phase in order to optimize the separability of
categories. This training procedure is described in section 4.

There exists already various ways to characterize the dimensionality
at different scales and to represent multiscale relations. For example
the fractal dimension \cite{fractal_dim_landscape} and the multifractal
analysis \cite{multifractal}. However these are not satisfying for
our needs. The fractal dimension is a single value that synthesize
the local space-filling properties of the point cloud over several
scales. It does not match the intuitive idea presented above in which
we aim at a signature of how the cloud dimension evolves over multiple
scales. The multifractal analysis synthesize in a spectrum how a signal
statistical moments defined at each scale relate to each other using
exponential fits (see \cite{multifractal} for more precise definitions,
we only give the main idea here as this is not the main topic of this
article). Unfortunately the multifractal spectrum does not offer a
discriminative power at any given scale, almost by definition (i.e.
it uses fits over multiple scales). Our goal is to have features defined
at each scale and then use a training procedure to define which combination
of scales allows to best separate two or more categories (such as
ground or vegetation). Some degree of classification is likely possible
using the aforementioned fractal analysis tools, but our new technique
is more intuitive and arguably better suited for the natural scenes
we consider. In the following we describe how the multi-scale dimensionality
feature is defined using the example of the simple pioneer salt marsh
environment in which only 2 classes exists : riparian vegetation (forming
individual patches) and ground (fine sand) (\ref{fig:Density-plots}).
More complex 3D multiclass cases (as in fig. \ref{fig:Otira-river-bed})
are addressed in section \ref{sub:3D-multiscale-multiclass}.

\subsection{\label{sub:Local-dimensionality-at}Local dimensionality at a given
scale}

\begin{figure}
\begin{centering}
\includegraphics[width=1\columnwidth]{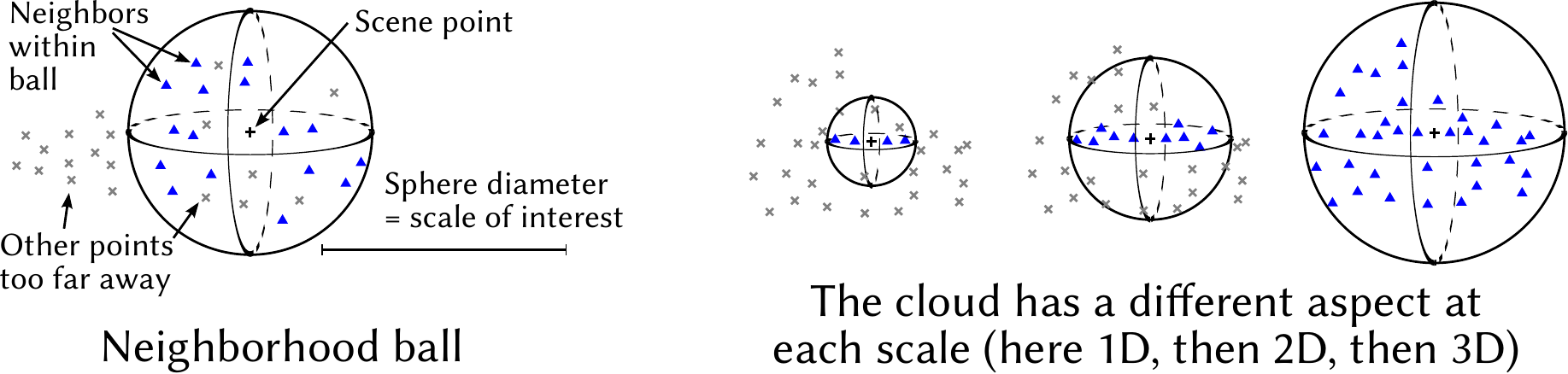}
\par\end{centering}

\caption{\label{fig:Neighborhood-ball}Neighborhood ball at different scales.
In this representation, outside points (gray stars) can be on the
side but also behind the neighborhood ball.}
\end{figure}

Let $C=\{P_{i}=(x_{i},y_{i},z_{i})\}_{i=1\ldots N}$ be a 3D point
cloud. The scale $s$ is here defined as the diameter of a ball centered
on a point of interest, as shown in Fig. \ref{fig:Neighborhood-ball}.
For each point in the scene the neighborhood ball is computed at each
scale of interest, and a Principal Component Analysis (PCA) \cite{stats_book}
is performed on the re-centered Cartesian coordinates of the points
in that ball.

\begin{figure}
\begin{centering}
\includegraphics[width=1\columnwidth]{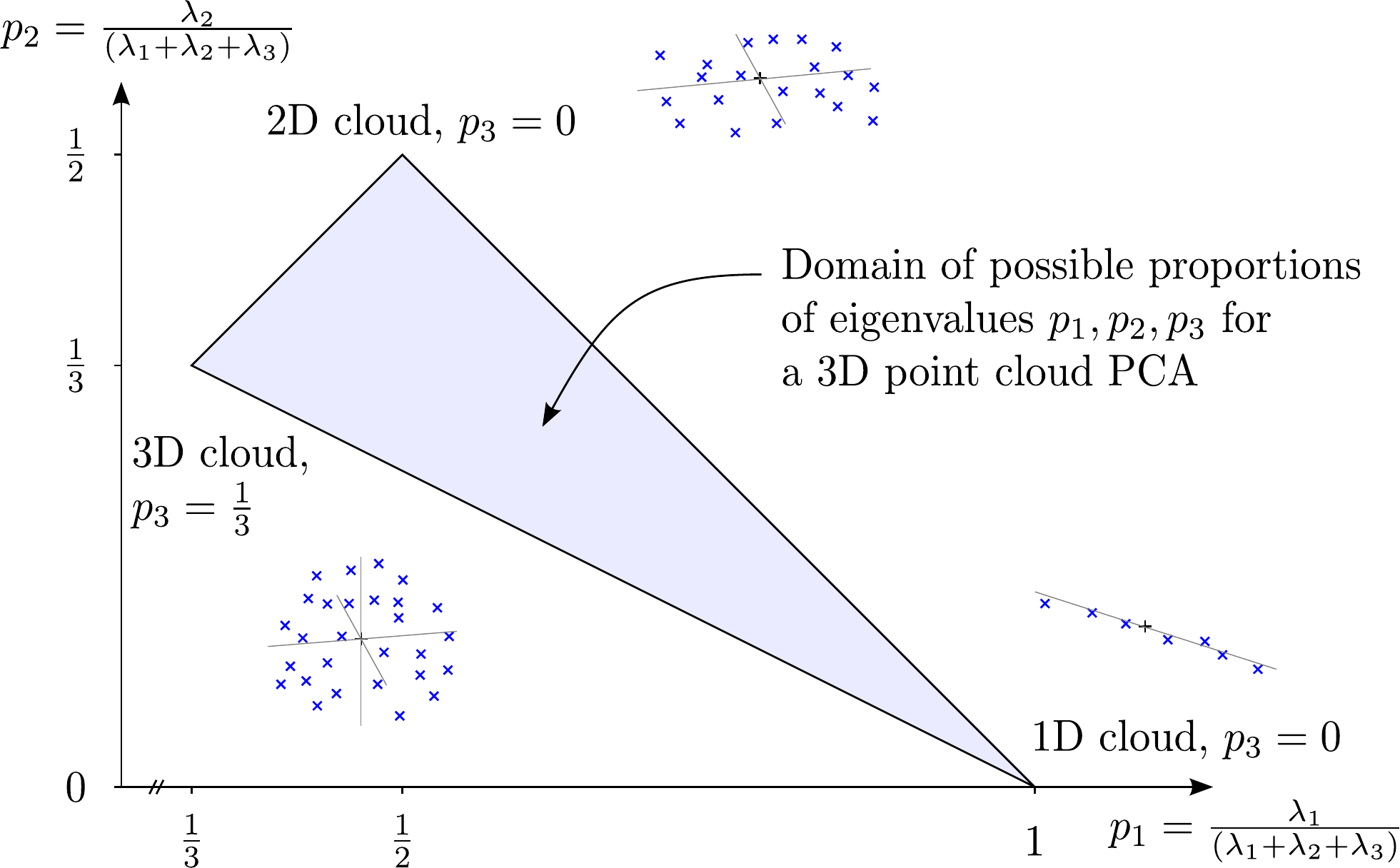}
\par\end{centering}

\caption{Representing the eigenvalues repartitions for the local neighborhood
PCA in a triangle.\label{fig:PCA-eigenvalues}}
\end{figure}

Let $\lambda_{i},i=1\ldots3$ be the eigenvalues resulting from the
PCA, ordered by decreasing magnitude: $\lambda_{1}\geq\lambda_{2}\geq\lambda_{3}$.
The proportion of variance explained by each eigenvalue is $p_{i}=\frac{\lambda_{i}}{\lambda_{1}+\lambda_{2}+\lambda_{3}}$.
Fig. \ref{fig:PCA-eigenvalues} shows the domain of all possible proportions.

When only a single eigenvalue $\lambda_{1}$ accounts for the total
variance in the neighborhood ball the points are distributed only
in one dimension around the reference scene point. When two eigenvalues
are necessary to account for the variance but the third one does not
contribute the cloud is locally mostly two-dimensional. Similarly
a fully 3D cloud is one where all three eigenvalues have the same
magnitude. The proportions of eigenvalues thus define a measure of
how much 1D, 2D or 3D the cloud appears locally at a given scale (see
Figs. \ref{fig:Neighborhood-ball} and \ref{fig:PCA-eigenvalues}).
Specifying these proportions is equivalent to placing a point $X$
within the triangle domain in Fig. \ref{fig:PCA-eigenvalues}, which
can be done using barycentric coordinate independently of the triangle
shape. Given the constraint $p_{1}+p_{2}+p_{3}=1$, a two-parameter
feature for quantifying how 1D/2D/3D the cloud appears can be defined
at any given point and scale.

A related measure has been previously introduced for natural terrain
analysis in the context of ground robot navigation \cite{Vandapel04,Lalonde06}
and urban lidar classification \cite{classifying_urban_lidar}. In
these applications, the eigenvalues of the PCA are used only as ratios
that are compared to three thresholds in order to define the feature
space. In the present study we not only consider the full triangle
of all possible eigenvalue proportions, as shown in \ref{fig:PCA-eigenvalues},
but also span the feature over multiple scales. The ``tensor voting''
technique from computer vision research predates our work in its use
of eigenvalues to quantify the dimensionality of the lidar data cloud
\cite{Schuster_2004,Kim_Medioni_2011}, although with a different
algorithmic approach. Our work is to our best knowledge the first
to combine the local dimensionality characterization over multiple
scales%
\footnote{We thank the editor for these references and remarks on our work.%
}. We chose PCA as it is a simple and standard tool for finding relevant
directions in the neighborhood ball \cite{stats_book}. Other projections
techniques (e.g. non-linear) could certainly be used for defining
different descriptors of the neighborhood ball geometry, but our results
below show that PCA is good enough already.

\subsection{Multiple scales feature}

\begin{figure}[!th]
\begin{centering}
\includegraphics[width=1\columnwidth]{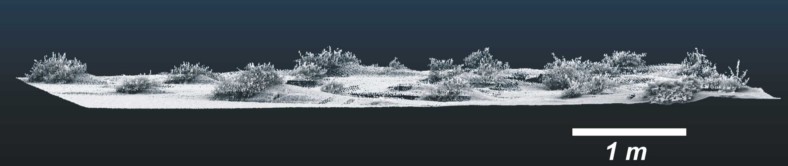}
\par\end{centering}

\begin{centering}
\includegraphics[width=1\columnwidth]{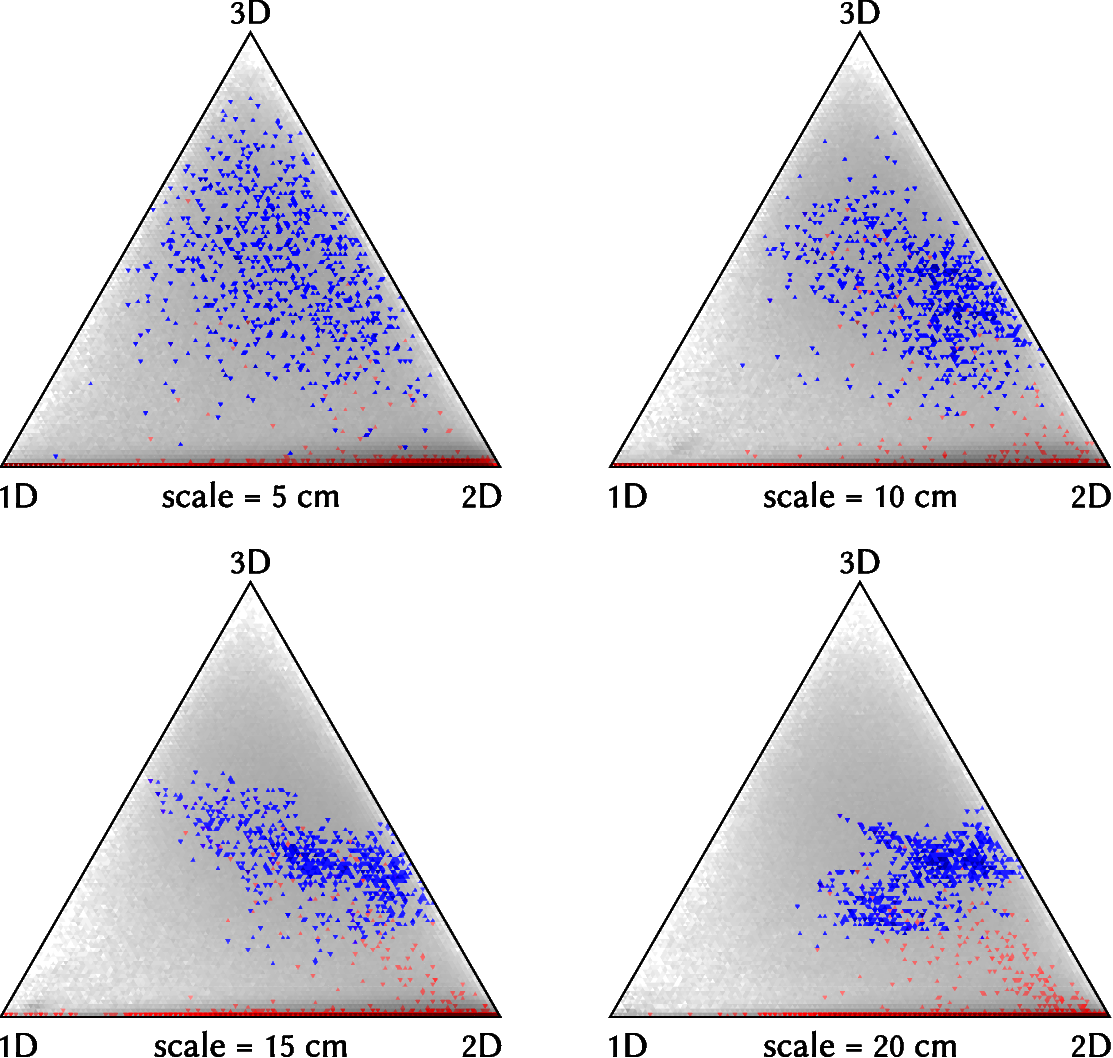}
\par\end{centering}

\begin{centering}
{\small Top : excerpt from a point cloud acquired in the Mont Saint-Michel
bay salt marshes (Fr), in a zone of pioneer riparian vegetation and
sand (point spacing from 2.3 to 14 mm). Bottom (with color available
online): Dimensionality density diagrams for one vegetation patch
(blue, appearing as dark gray when printed as gray), a patch of ground
(red, appearing as dark gray on the triangles bottom right 2D region),
and all other points of the scene (light gray). Each triangle is a
linearly transformed version of the space in Fig. \ref{fig:PCA-eigenvalues}
at the indicated scale. Each corner thus represents the tendency of
the cloud to be respectively 1D, 2D, or 3D.}
\par\end{centering}{\small \par}

\caption{Density plots of a scene represented in the proposed feature space
at different scales.\label{fig:Density-plots}}
\end{figure}

The treatment described in the previous section is repeated at each
scale of interest (see Fig. \ref{fig:Neighborhood-ball}). Given $N_{s}$
scales, we thus get for each point in the scene a feature vector with
$2.N_{s}$ values. This vector describes the local dimensionality
characteristics of the cloud around that point at multiple scales.
In the context of ground based lidar data there may be missing scales,
especially the smallest ones, because of reduced point density, nearby
shadows or scene boundaries. In that case the geometric properties
of the closest available larger scale is propagated to the missing
one in order to complete the $2.N_{s}$ values. Fig. \ref{fig:Density-plots}
shows an example of how a scene appears using this representation
for 4 scales.

Note how a single patch of vegetation (in blue in Fig. \ref{fig:Density-plots})
defines a changing pattern at different scales, but remains separated
from the ground (in red), hinting at a classification possibility.
However the rest of the scene (unlabeled, gray points) is spread through
the whole triangle at each scale: there is no clear cut between vegetation
and ground at any given scale. The solution is brought by \emph{considering
the multiscale vector in its entirety}, as a high-dimensional description,
and not as a succession of 2D spaces. This is described in the next
section.

\section{Classification}

The general idea behind the classification procedure is to define
the best combination of scales at which the dimensionality is measured,
that allows the maximum separability of two or more categories. Practically,
the user could have an intuitive sense of the range of scales at which
the categories will be the most geometrically different, but in many
cases, because of natural variability in shape and size of objects,
this is not a trivial exercise. We thus rely on an automated construction
of a classifier that finds the best combination of scales (i.e. all
scales contribute to the final classification but with different weights)
that maximizes the separability of two categories that the user has
previously manually defined (i.e. samples of vegetation and samples
of ground segmented from the point cloud). In the following we describe
the construction of the classifier and then present in section 5 typical
classification results and step-by-step application to natural data
sets.

\subsection{\label{sub:Probabilistic-classifier-in}Probabilistic classifier
in the plane of maximal separability}

\begin{figure}
\begin{centering}
\includegraphics[width=1\columnwidth]{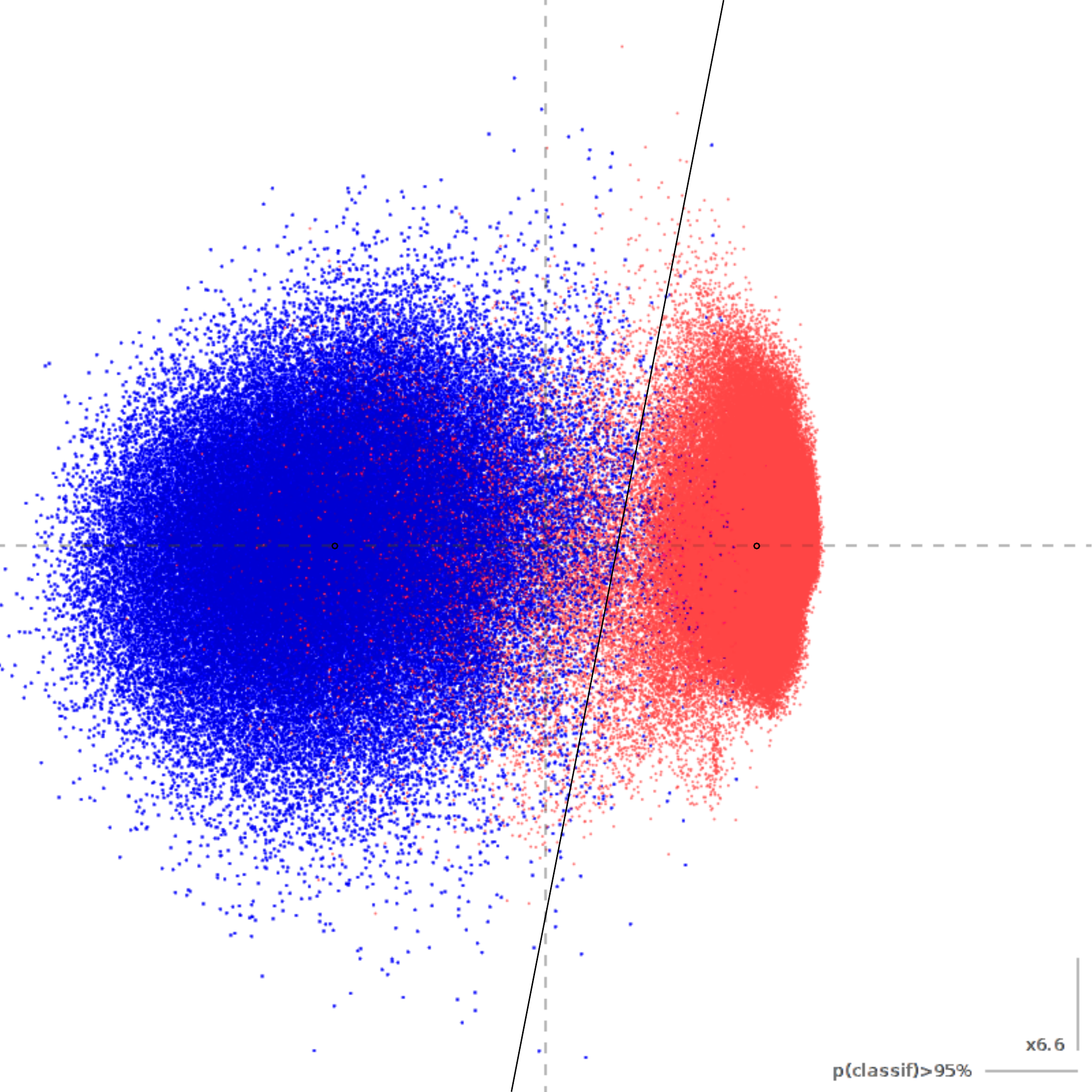}
\par\end{centering}

{\small Color is available online. Blue (dark gray): vegetation samples.
Red (light gray): soil. The classifier was obtained automatically
with a linear SVM using the process described in Section \ref{sub:Probabilistic-classifier-in}
in order to classify the benchmark described in Section \ref{sub:Quantitative-benchmark}.
The confidence level is given for the horizontal axis. The scaling
for the Y axis has no impact on the automated classification performance
but offers a better visualization, which is especially useful when
the user wishes to modify this file graphically.}{\small \par}

\caption{\label{fig:plane_separability}Classifier definition in the plane
of maximal separability.}
\end{figure}

The full feature space of dimension $2.N_{s}$ is now considered in
order to define a classifier that takes advantage of working simultaneously
on the data representation at multiple scales. This classifier is
defined in two steps: 1. by projecting the data in a plane of maximal
separability; and 2. by separating the classes in that plane. The
main advantage of processing this way is to get an easy supervision
of the classification process. Visual inspection of the classifier
in the plane of maximal separability is very intuitive, which in turn
allows for an easy improvement of the classifier if needed (e.g. changing
the separation line in Fig. \ref{fig:plane_separability} to make
a non-linear classifier)%
\footnote{Human intervention at this point allows for a powerful pattern recognition
beyond the capacities of the simple classifiers presented here. Moreover
some practical applications may require imbalanced accuracies for
each class. For example one may prefer to increase the confidence
in removing all the vegetation at the expanse of loosing a few data
points of ground. Allowing easy user intervention by means of a graphically
tunable classifier in a 2D plane of maximal separability nicely offers
these two advantages: improved pattern recognition and adaptability.
Automated processing is of course also possible and in fact forms
the default classifier on which the user can intervene if so desired.%
}.

The plane of maximal class separability is intuitively like a PCA
where only the 2 main components are kept, except that it optimizes
a class separability criterion instead of maximizing the projected
variance as the PCA would do. In principle any classifier allowing
a projection on a subspace can be used in an iterative procedure (including
non-linear classifiers with the kernel trick, see \cite{Maszczyk_Duch_2008}).
In the present work two linear classifiers are considered: Discriminant
Analysis \cite{Theodoridis_Koutroumbas_08} and Support Vector Machines
\cite{svm}. The rational is to assert the usefulness of our new feature
for discriminating classes of natural objects. Comparing the results
obtained with these two widely used linear classifiers validates that
the newly introduced feature does not depend on a complex statistical
machinery to be useful. We stress that last point: using one or the
other of these classifiers has little impact in practice (see the
results in section \ref{sub:Quantitative-benchmark}), but we had
to demonstrate this is actually the case and that using a simple linear
classifier is good enough for our use.

Let $F=\left\{ X=(x_{0},y_{0},x_{1},y_{1},\ldots,x_{Ns},y_{Ns})\right\} $
be the multiscale feature space of dimension $2.N_{s}$, with $(x_{i},y_{i})$
the coordinates within each triangle in Fig. \ref{fig:Density-plots}.
Consider the set of points $F^{+}$ and $F^{-}$ labeled respectively
by $+1$ or $-1$ for the two classes to discriminate (ex: vegetation
against ground). A linear classifier proposes one solution in the
form of an hyperplane of $F$ that best separates $F^{+}$ from $F^{-}$.
That hyperplane is defined by $w^{T}X-b=0$ with $w$ a weight vector
and $b$ a bias:
\begin{itemize}
\item Linear Discriminant Analysis proposes to set $w=\left(\Sigma_{1}+\Sigma_{2}\right)^{-1}\left(\mu_{2}-\mu_{1}\right)$
where $\Sigma_{c}$ and $\mu_{c}$ are the covariance matrix and the
mean vector of the samples in class $c$.
\item Support Vector Machines set $w$ so as to maximize the distance to
the separating hyperplane for the nearest samples in each class. The
Pegasos approach described in \cite{pegasos,dlib_svm} is used here
to compute $w$ since it is adapted to cases with large number of
samples while retaining a good accuracy.
\end{itemize}
In each case the bias $b$ is defined using the approach described
in \cite{Platt_1999}, which gives a probabilistic interpretation
of the classification: the distance $d$ of a sample to the hyperplane
corresponds to a classification confidence, internally estimated by
fitting the logistic function $p(d)=\frac{1}{1+exp(-\alpha d)}$.

The feature space $F$ is then projected on the hyperplane using $w$
and $b$, and the distance to the hyperplane $d_{1}=w_{1}^{T}X-b_{1}$
is calculated for each point. The process is repeated in order to
get the second-best direction orthogonal to the first, together with
the second distance $d_{2}$. The couple $(d_{1},d_{2})$ is then
used as coordinates defining the \textit{2D plane of maximal separability}.
Since there is a degree of freedom in choosing $w,b$ such that $w^{T}X-b=0$,
both axis can be rescaled such that $\alpha=1$. Thus the coordinates
$(d_{1},d_{2})$ in the separability plane are now consistent in classification
accuracy%
\footnote{To our knowledge this way of defining a 2D visualization in a plane
of maximal separability, while retaining an interpretation of the
scales in that plane using confidence values, is an original contribution
of this work.%
}. This consistency allows some post-processing in the plane. With
the current definition most classifiers would squash the data toward
the $X=Y$ diagonal%
\footnote{To see why, imagine the data being projected on the X axis with negative
coordinates for class 1 and positive coordinates for class 2. The
Y axis (second projection direction) also projects class 1/2 to negative/positive
coordinates. Hence the data is mostly concentrated along the diagonal.%
}. The post-processing consists in rotating the plane so that the class
centers are aligned on X, and then scaling the Y axis so the classes
have the same variance on average in both direction. This last step
is completely neutral with respect to the automated classifier that
draws a line in the plane (the optimal line could be defined whatever
the last rotation and scaling). However it is now much easier to visually
discern patterns within each class in the new rotated and rescaled
space, as can be seen is Fig. \ref{fig:plane_separability}.
That figure shows an example of classifier automatically obtained
using the data presented in Section \ref{sub:Quantitative-benchmark}.
The given scale of 95\% classification confidence is valid along the
X axis and the corresponding factor for the Y axis is indicated.

\subsection{\label{sub:Semi-supervised-learning}Semi-supervised learning}

One goal in developing this classification method was to minimize
user input (i.e. manually extracting and labeling data in the scene
is cumbersome) while maximizing the generalization ability of the
classifier . This is achieved by semi-supervised learning: using the
information which is present in the unlabeled points. The plane of
maximal separability is necessarily computed only with the labeled
examples. We search for a direction in this plane which minimizes
the density of all points along that direction (labeled and unlabeled),
while still separating the labeled examples. The assumption is that
the classes form clusters in the projected space, so minimizing the
density of unlabeled points should find these clusters boundaries.
When no additional unlabeled data are present the classes are separated
simply with a line splitting both with equal probability.

For a multi-class scenario (see Section \ref{sub:3D-multiscale-multiclass})
the final classifier is a combination of elementary binary classifiers.
In that case it may be that some cluster in the unlabeled data corresponds
to another class than the two being classified, which would fool the
aforementioned density minimization. A workaround is to use only the
labeled examples, or to rely on human visual recognition to separate
the clusters manually.

More generally the ability to visualize and keep control of the process
(this is not a ``black box'' tool) allows to tap on human perception
to better separate classes. But the ability to fully automate the
operations is retained, which is especially useful for large batch
processing. The user can always review the classifier if needed.

We developed a tool usable by non-specialists: the classifier is provided
in the form of a simple graphics file that the user can edit with
any generic, commonly available SVG editor%
\footnote{For example Inkscape, available at \url{http://www.inkscape.org/}
(as of 2012/01/19)%
}. The decision boundary can be graphically modified, thus quickly
defining a very powerful classifier with minimal user input. This
step is fully optional and the default classifier can of course be
taken without modification.

\subsection{Optimization}

The most time-consuming parts of the algorithm are computing the local
neighborhoods in the point cloud at different scales in order to apply
the local PCA transform (see Section \ref{sub:Probabilistic-classifier-in}),
as well as the SVM training process (computing the Linear Discriminant
Analysis is fast and not an issue, although even when using a SVM,
training a classifier is only needed once per type of natural environment).
We address these issues by allowing to compute the multiscale feature
on a sub-sampling of the scene called \textit{core points}. The whole
scene data is still considered for the neighborhood geometrical characteristics,
but that computation is performed only at the given core points.

This is a natural way of proceeding for lidar data: given the inhomogeneous
density there is little interest in computing the multiscale feature
at each point in the densest zones. A spatially homogeneous density
of core points is generally sufficient and allows an easier scene
manipulation and visualization%
\footnote{Both spatially homogeneous sub-sampling and scene manipulation are
easy to perform with free softwares like \href{http://www.danielgm.net/cc/}{CloudCompare}
\cite{CloudCompare}.%
}. However the extra data available in the densest zones is still used
for the PCA operation, which results in increased precision compared
to far away zones with less data points. We also preserve the local
density information and the classification confidence around each
core point as a measure of that precision. When classifying the whole
scene, each scene point is then given the class of the nearest core
point.

As a result the user is offered a trade-off between computation time
and spatial resolution : it is possible to call the algorithm on the
whole scene (each point is a core point) or to call the algorithm
on a sub-sampling of the user choice (e.g., an homogeneous spatial
density of core points).

\section{Results}

\subsection{\label{sub:Quantitative-benchmark}Quantitative benchmark on ground
and riparian vegetation classification}

In order to quantitatively assess the performance of the classifier,
examples were selected from the pioneer salt marsh scene (see Fig.
\ref{fig:Density-plots} for an excerpt of this scene) in which two
classes can be defined : riparian vegetation and ground. These examples
represent various vegetation patch sizes and shapes, shadow zones,
flat ground, small ripples, data density changes and multiple scanner
positioning. The data set comprises approximately 640000 points, manually
classified into 200000 belonging to vegetation and 440000 for ground.
This data set is provided online together with the software (link
given at the end of this paper) so it can be reused for comparative
benchmarks.

The classifier is trained to recognize vegetation from ground in the
first set of examples, using about half of the aforementioned data.
Its performance is then assessed on a the remaining half of the data
that was not used for training. This is not only the standard procedure
in the machine learning field (to detect when the algorithm learns
details of a particular data set that are not transposable to other
data sets, i.e. the over-fitting issue), but also what is expected
from our new technique. We aim at an excellent generalization ability:
the algorithm must be able to recognize the vegetation in unknown
scenes, not only just on the samples it was presented.

We use the balanced accuracy measure to quantify the performance of
the classifier in order to account for the different number of points
in each class. With $tv,\, tg,\, fv,\, fg$ the number of points truly($t)$/falsely($f$)
classified into the vegetation($v$)/ground($g$) classes, the balanced
accuracy is classically defined as $ba=\frac{1}{2}\left(a_{v}+a_{g}\right)$
with each class accuracy defined as $a_{v}=\frac{tv}{tv+fg}$ and
$a_{g}=\frac{tg}{tg+fv}$. We use the Fisher Discriminant Ratio $fdr$
\cite{Theodoridis_Koutroumbas_08} in order to assess the class separability.
The classifier assigns for each sample a signed distance $d$ to the
separation line, using negative values for one side and positive values
for the other. The measure of separability is defined as $fdr=\left(\mu_{2}-\mu_{1}\right)^{2}/\left(v_{1}+v_{2}\right)$
with $\mu_{c}$ and $v_{c}$ the mean and variance of the signed distance
$d$ for each class $c$. Note that the class separability could still
be high despite a mediocre accuracy (e.g., separation line positioned
on a single side from both classes). This would merely indicate a
bad training with potential for a better separation. Hence both $ba$
and $fdr$ are useful measures for asserting separately the role of
the classifier and the role of the newly introduced feature in the
final classification result. A large $ba$ value indicates a good
recognition rate ($ba=50\%$ indicates random class assignment) on
the given data set, and a large $fdr$ value indicates that classes
are well separated (an indication that the $ba$ score is robust).

\begin{table}
\begin{centering}
\begin{tabular}{|c|c|c|c|c|}
\hline 
\multicolumn{2}{|c|}{LDA classifier} & Accuracy & $ba$ & $fdr$\tabularnewline
\hline 
\multirow{2}{*}{Training set} & Vegetation & 98.3\% & \multirow{2}{*}{97.9\%} & \multirow{2}{*}{12.3}\tabularnewline
\cline{2-3} 
 & Ground & 97.6\% &  & \tabularnewline
\hline 
\multirow{2}{*}{Testing set} & Vegetation & 99.3\% & \multirow{2}{*}{97.6\%} & \multirow{2}{*}{11.0}\tabularnewline
\cline{2-3} 
 & Ground & 95.9\% &  & \tabularnewline
\hline 
\end{tabular}
\par\end{centering}

~

\begin{centering}
\begin{tabular}{|c|c|c|c|c|}
\hline 
\multicolumn{2}{|c|}{SVM classifier} & Accuracy & $ba$ & $fdr$\tabularnewline
\hline 
\multirow{2}{*}{Training set} & Vegetation & 98.7\% & \multirow{2}{*}{98.0\%} & \multirow{2}{*}{11.1}\tabularnewline
\cline{2-3} 
 & Ground & 97.3\% &  & \tabularnewline
\hline 
\multirow{2}{*}{Testing set} & Vegetation & 99.6\% & \multirow{2}{*}{97.5\%} & \multirow{2}{*}{11.0}\tabularnewline
\cline{2-3} 
 & Ground & 95.4\% &  & \tabularnewline
\hline 
\end{tabular}
\par\end{centering}

{\small The performances of each classifier is measure using the Balanced
Accuracy (ba) and the Fisher Discriminant Ratio (fdr). Both are described
in the main text.}{\small \par}

\caption{Quantitative benchmark for separating vegetation from ground.\label{tab:benchres}}
\end{table}

Table \ref{tab:benchres} shows the results of the benchmark. The
classifier that was used is fully automated, without human intervention
on the decision boundary, and taking 19 scales between 2cm and 20cm
every cm (larger scales do not improve the classification, see Fig.
5 for the typical vegetation size $\approx40$cm). We used our software
default quality / computation time trade-off for the support vector
machine classifier training in order to adequately assess the results
of our algorithm in usual conditions. The algorithm was forced to
classify each point, while in practice the user may decide to ignore
points without enough confidence in the classification (see Section
\ref{sub:3D-multiscale-multiclass}). Nevertheless the balanced accuracy
that was obtained both on the training set and the testing set is
very good. This not only shows that the algorithm is able to recover
the manually selected vegetation/soil (train set accuracy) but that
it is able to generalize to terrain data it had not seen before. This
is of great importance for large campaigns: we can train the algorithm
once on a given type of data and then apply the classifiers to a large
quantity of further measurements without having to re-train the algorithm.
\begin{table*}
\begin{centering}
\begin{tabular}{|l|c|c|c|c|c|c|c|c|}
\hline 
\multirow{2}{*}{LDA and SVM} & \multicolumn{2}{c|}{5cm} & \multicolumn{2}{c|}{10cm} & \multicolumn{2}{c|}{15cm} & \multicolumn{2}{c|}{20cm}\tabularnewline
\cline{2-9} 
 & $ba$ & $fdr$ & $ba$ & $fdr$ & $ba$ & $fdr$ & $ba$ & $fdr$\tabularnewline
\hline 
Training & 97.0\% & 5.2 & 97.1\% & 6.5 & 96.6\% & 5.6 & 95.7\% & 4.6\tabularnewline
\hline 
Testing & 97.3\% & 6.4 & 96.9\% & 6.5 & 95.7\% & 4.8 & 94.1\% & 3.7\tabularnewline
\hline 
\end{tabular}
\par\end{centering}

{\small The results for both classifiers differ only at the fourth
digit for the Balanced Accuracy ($ba$) and at the third for the Fisher
Discriminant Ratio ($fdr$), so the tables were merged.}{\small \par}

\caption{\label{tab:Single-scale-benchmark}Single scale benchmark results
at selected scales}
\end{table*}

Table \ref{tab:Single-scale-benchmark} shows the result of the classification
using single scales only. The advantage of using a multi-scale classifier
is apparent: it offers a better accuracy than any single scales alone.
The difference is more pronounced for the discriminative power, with
the multi-scale classifier offering almost twice as much class separability.
Although this is the expected behavior, some classifiers are sensitive
to noise and adding scales with no information would potentially decrease
the multi-scale performance. The scales from 2cm to 20cm not shown
in Table \ref{tab:Single-scale-benchmark} have similar properties
and performance levels, with slightly better results for single scales
between 5 and 10 cm. Even with this observed performance peak there
is no characteristic scale in this system as discriminative information
is present at all scales: the point of the multi-scale classifier
is precisely to exploit that information.

In this example, both classifiers (LDA and SVM) give the same results
at each scale, and are equally suitable for the multiscale feature
(Table \ref{tab:benchres}). In other scenarios the situation might
be different, but overall this confirm our method does not need a
complicated statistical machinery (like the SVM) for being effective,
and using a simple linear classifier (like the LDA) is good enough.
In any case we achieve at least 97.5\% classification accuracy.

\begin{figure*}[!]
\includegraphics[width=1\textwidth]{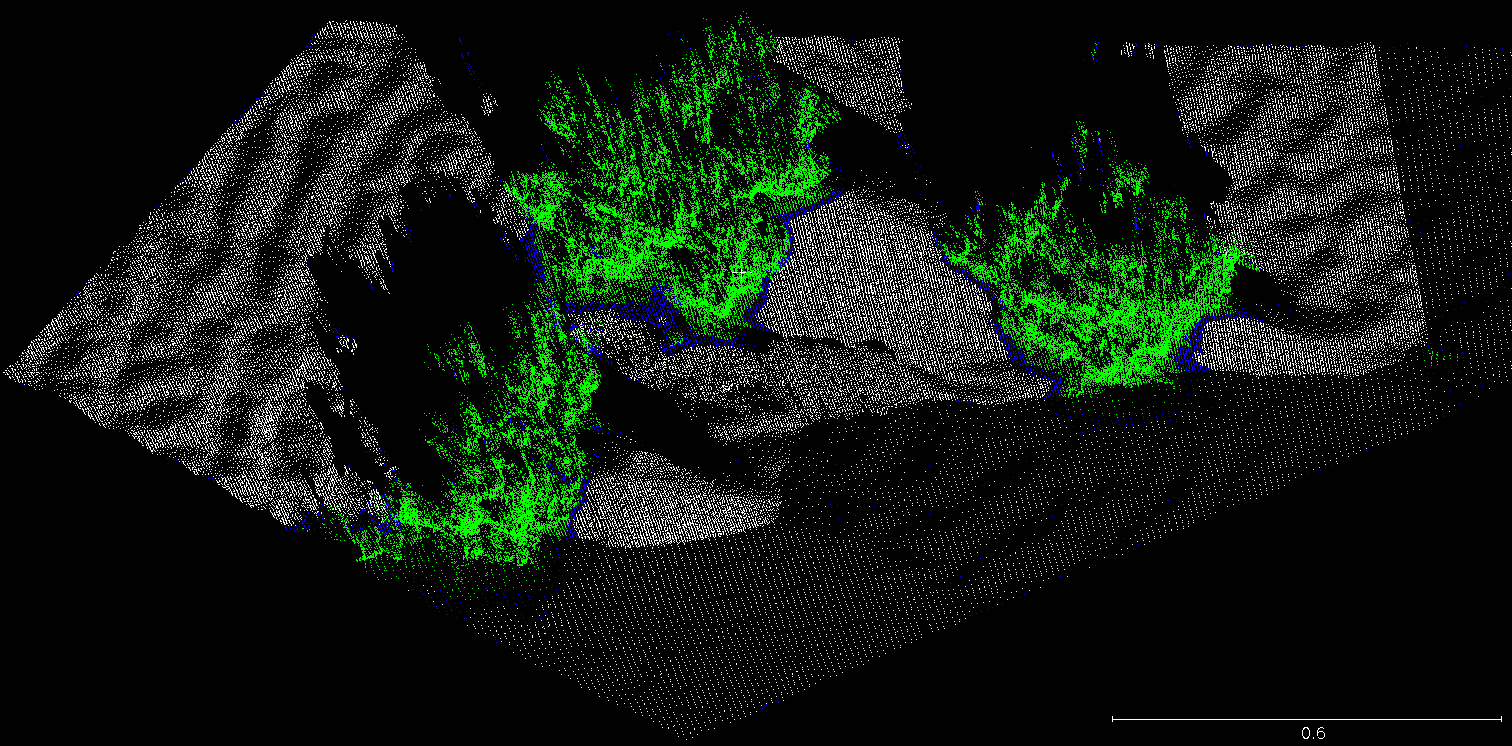}

{\small Color is available online. White: Points recognized as ground.
Green (light gray): Points recognized as vegetation. Blue(dark gray):~Points
for which the confidence in the classification is less than 80\%.
Scale is in meters.}{\small \par}

\caption{Excerpt of the quantitative benchmark test set classification\label{fig:bench_visu}}
\end{figure*}

Figure \ref{fig:bench_visu} visually shows the result of the classification
on a subset of the test data using the multi-scale SVM classifier
obtained with the fully automated procedure. Points with a low classification
confidence are highlighted in blue. They correspond mostly to the
boundary between ground and vegetation. Figure \ref{fig:bench_visu}
shows that the algorithm copes very well with the irregular density
of points, the shadow zones and the ripples. The actual classifier
definition is shown in Fig. \ref{fig:plane_separability}. 

\subsection{\label{sub:3D-multiscale-multiclass}3D multiscale classifiers with
multiple classes}

\subsubsection{Dealing with multiple class}

Combining multiple binary classifiers into a single one for handling
multiple classes is a longstanding problem in machine learning \cite{multiclass}.
Typically the problem is handled by training ``one against one''
(or ``one against others'') elementary binary classifiers, which
are then combined by a majority rule. This is what the automated tool
suite CANUPO proposes in the present context, following the common
practice in the domain.

Additional extensions are of course possible in future works. Recent
developments on advanced statistical techniques \cite{multiclass}
deal with the issue of training and then combining the elementary
classifiers. However in the present context we wish to retain a possible
intervention on the classifiers using a graphical editor. Moreover
context-dependent choices like favoring one class over the other need
to be allowed. It may thus be more efficient to separate classes one
by one and combine the results, as is explained in the next section.

\subsubsection{Application to a complex natural environment}

\begin{figure*}
\includegraphics[width=1\textwidth]{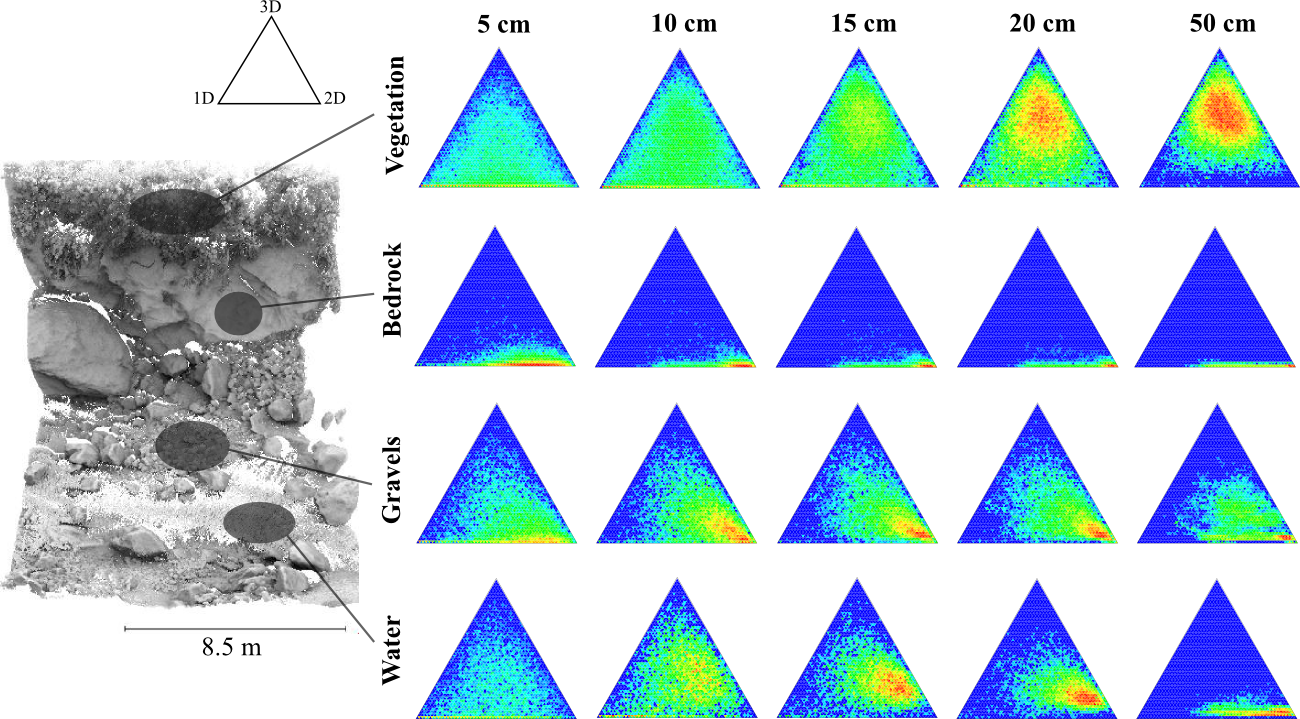}

\caption{Multiscale, multi-class scenario}
{\small \label{Fig:Diag_ternaire_OTIRA}Left : excerpt from the point
cloud of fig. \ref{fig:Otira-river-bed}. Right: dimensionality diagrams
for the four main classes of a mountain river environment at scales
ranging from 5 to 50 cm. Colors from blue to red correspond to the
density of points from the training dataset and characterize the degree
of clustering around a given dimensionality.}
\end{figure*}

In the following we illustrate the capabilities of the method in classifying
complex 3D natural scenes. A subset of the Otira River scene (fig.
\ref{fig:Otira-river-bed}) was chosen, and four main classes defined:
vegetation, bedrock surface (on the channel bank and large blocs),
gravel surfaces and water. Figure \ref{Fig:Diag_ternaire_OTIRA} presents
the dimensionality density diagrams of one training patch for each
class and scales ranging from 5 to 50 cm. As intuitively expected,
vegetation is mostly 1D and 2D at small scale (leaves, stem) and becomes
dominantly 3D at scales larger than 15 cm. However, the clustering
of points is only significant at scales larger than 20 cm. Bedrock
surfaces are mostly 2D at all scales, with some 1D-2D features occurring
at fine scales corresponding to fractures. Gravel surfaces exhibit
a larger scatter at all scales owing to the large heterogeneity in
grain sizes. The 3D component is more important at intermediate scales
(10 to 20 cm) than at small or very large scales. This illustrates
the transition from a scale of analysis smaller than the dominant
gravel size (i.e., gravels appears as dominantly 2D curved surfaces),
and then larger than an assemblage of gravels (i.e., gravel roughness
disappears). As explained in section \ref{sec:Study-sites-and}, whitewater
surface can be picked by the laser, whereas in general it does not
reflect on clear water \cite{Milan10}. Yet, even at small scale the
water does not appear purely 2D as the water surface is uneven and
the laser penetrates at different depth in the bubbly water surface.
Indeed, the signature is quite multidimensional for scales up to 20
cm, and only around 20 cm does the water surface appear to significantly
cluster along a 2D-3D dimensionality. At larger scale, the water becomes
significantly 2D.

The multi-scale properties of the various classes show that there
is not a single scale at which the classes could be distinguished
by their dimensionality. Vegetation and bedrock are quite distinct
at large scale, but bedrock, gravel and water are too similar at this
scale to be labeled with a high level of confidence. Only at smaller
scales (10-20 cm) can bedrock be distinguished from gravel and water.
This visualization also shows that gravel and water will be difficult
to distinguish owing to their very similar dimensionality across all
the scales. 

\begin{figure*}
\includegraphics[width=1\textwidth]{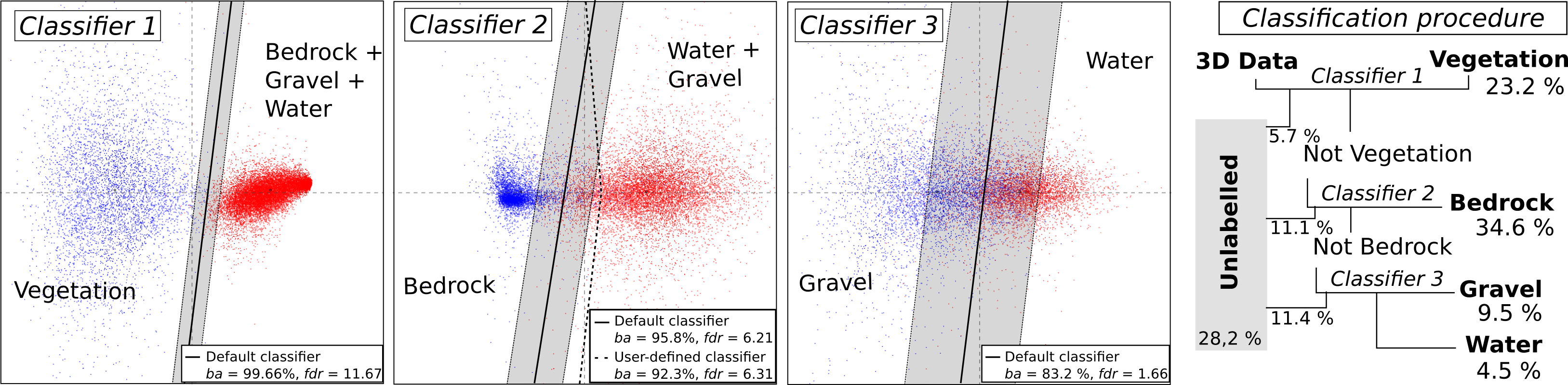}

\caption{Classification process and classifiers for the multiscale case}

{\small \label{Fig:Classifiers Otira}Semi-supervised classifiers
for the Otira river dataset and classification procedure diagram.
For each classifier, the gray area indicates the portion of space
in which the classification confidence is lower than 0.95. For Classifier
2, the manually modified supervised classifier is targeted to preserve
more systematically bedrock surfaces at the expense of non-bedrock
surfaces. In the classification procedure diagram, percents indicate
the proportion of each class in the supervised classification which
uses confidence intervals of 0.9 for Classifier 1, 0.8 for Classifier
2 and Classifier 3.}
\end{figure*}

\begin{figure*}
\begin{centering}
\includegraphics[width=1\textwidth]{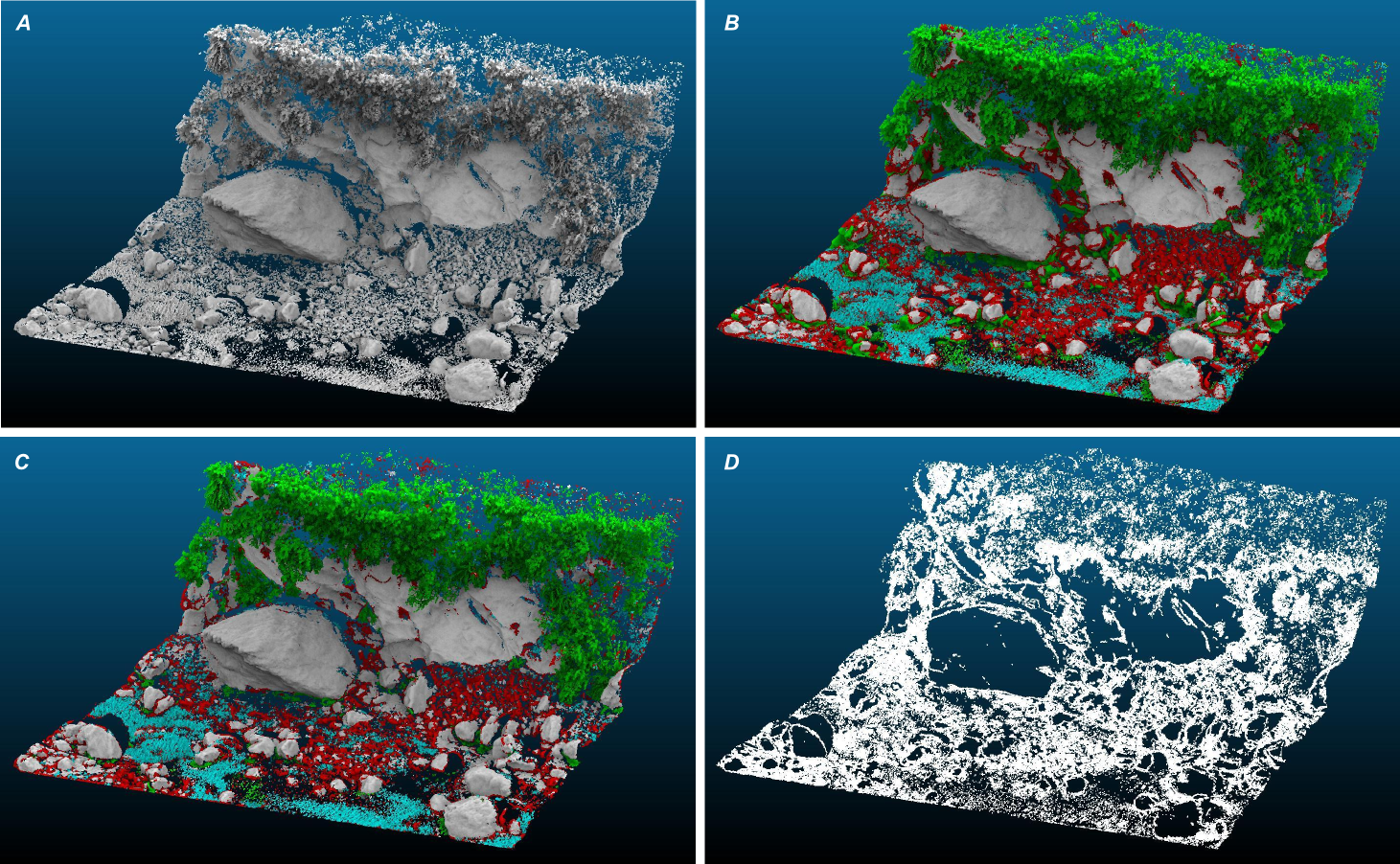}
\par\end{centering}

\caption{Result of the classification process for the mountain river dataset}

{\small \label{Fig:Classified Scene Otira}A: Original Otira River
scene (minimum point spacing = 1 cm). B: Default classification (green:
vegetation, gray: bedrock, red: gravel, blue: water) according to
the process described in fig. \ref{Fig:Classifiers Otira}. C : User-improved
classification. D : unlabeled points (28,2 \% of the total points).}
\end{figure*}

In the following, approximatively 5000 core points for each class
were selected for the training process. Their multiscale characteristics
were estimated using the complete scene rather than excerpts of the
class only. Points in the original scene have a minimal spacing of
1 cm corresponding to \textasciitilde{} 1.17 million points. The actual
classification operates on subset of 330000 core points with a minimum
spacing of 2 cm. The multi-classes labeling was achieved using a series
of 3 binary classifiers (fig. \ref{Fig:Classifiers Otira}) all using
the same set of 22 scales (from 2 cm to 1 m). An automated classification
(i.e., the only user interaction was in defining the classes and the
initial training sets) is presented, as well as examples of possible
user alterations.These alterations are of three types : changing the
initial training sets, modifying the classifier, and defining a classification
confidence interval. Given that users improvements depend on specific
scientific objectives (e.g., documenting vegetation, characterizing
grain sizes or measuring surface change), they cannot all be discussed
completely here. We present a case in which the classification of
bedrock surfaces was slightly optimized. The LDA approach was used
for all classifier definitions as the results were on par or slightly
better than a SVM approach. Figure \ref{Fig:Classified Scene Otira}
presents the results for the original data, the automated and the
user-improved classification results.

The first classifier separates vegetation from the three other classes.
The automated training procedure results in a \textit{$ba$} of 99.66
\% approaching perfect identification of vegetation on the training
sets. The very high level of separability is reflected by a large
\textit{$fdr$} value (11.67) and a very small classification uncertainty
in the projected space (fig. \ref{Fig:Classifiers Otira}). As shown
in figure \ref{Fig:Classified Scene Otira}, the automated classification
of vegetation is excellent with very limited false positives appearing
in overhanging parts of large blocs where the local geometry exhibits
a dimensionality across various scales too similar to vegetation.
The precision of the labeling is also excellent as small parts of
bedrock between or behind vegetation are correctly identified, and
small shrubs are correctly isolated amongst bedrock surfaces. Nevertheless
it is still possible to improve this classifier by using the incorrectly
classified overhanging blocs in the training process (5000 core points
were added). This 5 minutes operation results in a better handling
of false vegetation positive, and retains excellent characteristics
on the original training sets (\textit{$ba$} = 98.2 \%\textit{, $fdr$
=} 9.89). A classification confidence interval of 90 \% was also set
visually in the CloudCompare software \cite{CloudCompare} by displaying
the uncertainty level of each core point and defining the optimum
between quality and coverage of the classification. This left aside
5.7 \% of the original scene points unlabeled. 

Classifier 2 separates bedrock surfaces from water and gravel surfaces
(fig. \ref{Fig:Classifiers Otira}). The automated training procedure
lead to a \textit{$ba$} of 95.7 \% and \textit{$fdr$} of 6.21 on
the original training sets. Because gravels exhibits a wide range
of scales from pebbles to boulders, it is not possible to fully separate
the bedrock and gravel classes as the largest gravels tend be defined
as rock surfaces. Fracture and sharp edges of blocs tend to be classified
as non-bedrock as they are 3D feature at small scale and 2D as large
scale (as is gravel). Yet, as in the previous case, the confidence
level defined at 0.95 remains small compared to the size of the two
clusters in the projected space. While the original classifier was
already quite satisfactory, it was tuned to primarily isolate rock
surfaces by changing manually the classifier position in the hyperplane
projection (\textit{$ba$} = 92.3 \%\textit{, $fdr$ =} 6.31 ,fig.
\ref{Fig:Classifiers Otira}). A classification confidence interval
of 80 \% was also used which left 17 \% of the remaining points unlabeled
(fig. \ref{Fig:Classified Scene Otira}).

Classifier 3 separates water from gravel surfaces (classifier 3, fig.
\ref{Fig:Classifiers Otira}).The automated training procedure lead
to a ba of 83,2 \%. As expected from the similarity of the dimensionality
density diagrams (fig. \ref{Fig:Diag_ternaire_OTIRA}), the two classes
are more difficult to separate than in the previous two classifications
and the confidence level defined at 0.95 overlaps significantly the
two classes. Yet, figure \ref{Fig:Classified Scene Otira} shows that
the classifier manages to correctly label the whitewater and gravel
surfaces corresponding to non-trained datasets. Being quite effective,
the default classifier was not altered. A confidence interval of 80
\% was used resulting in 78 \% of the remaining points being labeled. 

The end result of this process is a 3D scene (fig. \ref{Fig:Classified Scene Otira}).
As shown in fig. \ref{Fig:Classified Scene Otira} the default parameters
already give an excellent first order classification. The fine-tuning
previously described do not represent the best classification possible,
but rather an example of how the automated approach can be rapidly
tweaked to give some improvement. On a practical note, the simplest
way to improve the default classifier in this example is to add in
the training process some of the false positive and false negative
results of a first training, rather than manually altering the classifier.
Defining a confidence level during the classification process is very
useful as the amount of data is so large that labeling only 70 \%
of the points is not detrimental to the interpretation of the results.
The classes can be further cleaned by removing isolated points using
the volumetric density of data calculated during the multi-scale analysis.

\subsection*{5.3 Single scale vs Multiple scale classification}

Table \ref{Table-single-multiple} presents the balanced accuracy
of the three classifiers used in the Otira river scene (fig. \ref{Fig:Classifiers Otira})
trained with the same subsets but using a single spatial scale (5,
10, 20, 50, 75 or 100 cm). For each classifier, the balanced accuracy
of the multiple scale classification is systematically better than
the single scale ones. The improvement is very significant for Classifier
3 (83.2 \% vs 70.9 \% for single class). Most importantly, the separability
of classes (as measured by the $fdr$) is always increased at least
two to three times for Classifiers 2 and 3 (and by 40 \% for Classifier
1). The increased separability is the key advantage of the multi-scale
approach. It allows a larger geometrical inhomogeneity within a given
class, and a better generalization of the results than a single scale
approach.

Compared to a single scale classification at 1 m, the improvement
of the multi-scale Classifier 1 (vegetation vs not vegetation) seems
more marginal. However, by comparing the classification results on
the Otira river data, the multi-scale classifier is more precise than
the single-scale case: small shrubs within bedrock, that are not correctly
classified by the single large scale classifier, are correctly retrieved
with the multi-scale approach. Similarly, incorrectly classified zones
below blocs for both classifiers are more extended with the single-scale
classifier. Therefore the multi-scale classification is qualitatively
improved, which is not reflected by the quantitative $ba$ measure
alone.

We conclude that the multi-scale analysis always improve significantly
the classification compared to a single scale analysis on one or all
of the following aspects : discrimination capacity, separability of
the classes and spatial resolution. 

\begin{table}
\begin{tabular}{|c|c|c|c|c|c|c|}
\hline 
Scale & \multicolumn{2}{c|}{Classifier 1} & \multicolumn{2}{c|}{Classifier 2} & \multicolumn{2}{c|}{Classifier 3}\tabularnewline
(cm) & \multicolumn{2}{c|}{vegetation} & \multicolumn{2}{c|}{bedrock} & \multicolumn{2}{c}{water,gravel}\tabularnewline
\cline{2-7} 
 & \textit{$ba$ \%} & \textit{$fdr$} & \textit{$ba$ \%} & \textit{$fdr$} & \textit{$ba$ \%} & \textit{$fdr$}\tabularnewline
\hline 
\hline 
2-100 & \textbf{99.66} & \textbf{11.67} & \textbf{95.7} & \textbf{6.21} & \textbf{83.2} & \textbf{1.66}\tabularnewline
\hline 
5 & 67.51 & 0.18 & 78.75 & 1.04 & 70.28 & 0.32\tabularnewline
\hline 
10 & 58.6 & 0.03 & 88.47 & 1.89 & 69.36 & 0.46\tabularnewline
\hline 
20 & 82.23 & 1.76 & 92.15 & 2.84 & 62.03 & 0.14\tabularnewline
\hline 
50 & 95.59 & 5.73 & 85.24 & 1.56 & 68.28 & 0.41\tabularnewline
\hline 
75 & 98.24 & 7.55 & 79.85 & 1.03 & 69.85 & 0.43\tabularnewline
\hline 
100 & 98.98 & 8.2 & 78.74 & 0.77 & 70.9 & 0.50\tabularnewline
\hline 
\end{tabular}

\caption{Comparison of Balanced Accuracy on Single scale and Multi-scale classifier}

{\small \label{Table-single-multiple}Results of semi-supervised LDA
classification for single and multiple scales using the original training
sets of the Otira river classifiers (fig. \ref{Fig:Classifiers Otira}).
Results are similar using an SVM approach.}
\end{table}

\begin{figure}[!t]
\includegraphics[width=1\columnwidth]{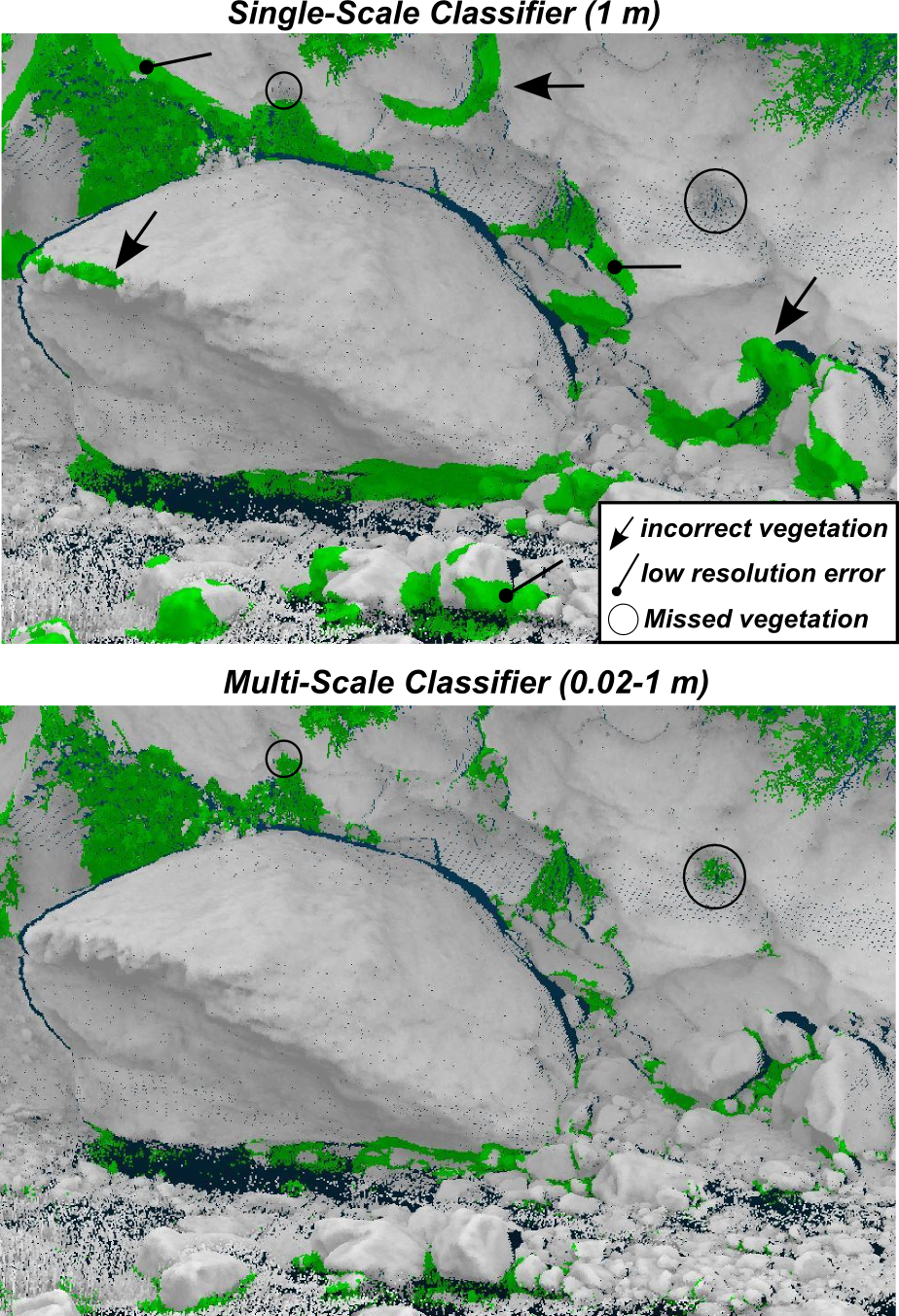}

\caption{Comparison of best single scale vegetation classifier with multi-scale
classifier}

{\small \label{Fig: single vs multiple scales} Classification results
for vegetation detection using a multi or single scale (1 m) classifier.
Even though the balanced accuracy is similar on the training set the
multi-scale classification is much more precise and less prone to
errors when generalized to the whole scene. The single scale classifier
misses small shrubs on the bedrock and incorrectly classify large
bloc borders as vegetation.}
\end{figure}

\section{Discussion / openings\label{sec:Discussion-/-openings}}

While many studies have focused on the classification of ground vs
vegetation, or buildings in 2.5D airborne lidar data using purely
geometrical approaches (e.g., \cite{BareEarth,Antonorakis08,classifying_urban_lidar}),
none can really apply to dense 3D point clouds obtained from ground
based (fixed or mobile) lidar data in which a fully 3D approach is
needed. Such 3D approach have been pursued using a dimensionality
measurement at a given scale \cite{Vandapel04,Lalonde06} to detect
1D (e.g. tree trunk or cables), 2D (ground) and 3D (vegetation) objects.
However, because natural surfaces exhibit a large range of characteristic
scales and natural objects within a given class can have a large degree
of geometric heterogeneity (i.e. vegetation or sediment), a single
scale can rarely classify an entire scene robustly. We have thus introduced
a multi-scale analysis of the local geometry of point clouds to cope
with the aforementioned issues, which exhibits good performance even
with simple linear classifiers. By doing so the selection of a specific
or characteristic scale is not needed. We have shown that the combination
of scales systematically improves the separability of classes compared
to a single scale analysis (Table \ref{Table-single-multiple}, Table
\ref{tab:benchres} vs Table \ref{tab:Single-scale-benchmark}), sometimes
dramatically. The multi-scale analysis also allows retention of a
detailed spatial resolution compared to a single large scale analysis.
To further account for the geometrical complexity of natural environments,
the user is free to set a level of confidence in the classification
process that will control the balance between confidence and coverage
of the classification. Given the large amount of data available in
TLS, it is often more interesting to not classify 30 \% of the data,
in order to keep 70 \% for which classes are correctly attributed.

Because all scales contribute to a varying degree to the classification
process, the method is relatively robust to shadow effects, missing
data and irregular point density (e.g. fig. \ref{Fig:Classified Scene Otira}b
and \ref{fig:bench_visu}): even if the dimensionality cannot be characterized
over a certain range of scales (e.g. small ones because of low point
density, large ones because of shadow effects), other scales are used
to classify a point, albeit with a smaller degree of confidence. Interestingly,
qualitative inspection of the classification results shows that obvious
spurious mixed points created at the edge of objects tend to be classified
with a low level of confidence (provided that the scene has a relatively
high point density and that the small scale dimensionality significantly
contribute to the classification). This is explained by the low point
density around mixed points (because no real object exists at their
location) and the resulting lack of a good dimensionality characterization
at small scale. Although systematically quantifying this effect if
out of the scope of this work, this observation suggests that using
a relatively high level of confidence during the classification process
helps in cleaning the resulting classes from spurious mixed-points.

We chose the dimensionality of the point cloud at a given scale as
a continuous measure of the local scene geometry. This is an intuitive
perception of the surface that can capture many aspects of natural
geometries (\cite{Vandapel04,Lalonde06}), in particular the dichotomy
between 3D vegetation and 2D surfaces. However, the multi-scale classification
could also use other geometrical measures depending on the final objectives
of the classification. Surface orientation, curvature, mathematical
derivatives of a local surface \cite{SupervisedClassif} or the degree
of conformity to a given geometry (sphere, cylinder ....e.g., \cite{Vosselman_3Dgeometry})
could also have been used in the construction of the classifier. The
surface angle with the vertical is already implemented in the classifier
but was not used here. It could be used to separate channel banks
from river bed for instance, or as an additional constraint to discriminate
the water surface from the gravels (which are rarely completely horizontal
compared to water). We note that for vegetation classification, the
dimensionality measurement is particularly effective and simple to
define for 3D point clouds. Indeed, even at a single well chosen scale,
the dimensionality criteria performs already well to detect vegetation
(i.e., fig. \ref{Fig: single vs multiple scales})(\cite{Vandapel04,Lalonde06}).

Each point captured by lidar (airborne or terrestrial) also comes
with a measure of the reflected laser intensity and in some cases
with optical imagery information (RGB) that could be used in the classification
process. Using the reflected laser intensity for classification purposes
has been attempted for airborne (e.g. \cite{lidar_reflectance_survey,ALidar_Icorrection})
and ground-based lidar \cite{Franceschi09,lidar_surface_reflectance,Lichti05,Guarnieri09,Correction_methods_intensity_2011}.
In this latter case, the difficulties are numerous as the reflected
intensity is a complex function of distance from the scanner, incidence
angle, and surface reflectance \cite{Correction_methods_intensity_2011,Lichti05}
(which on top of the physico-chemical characteristics of the material
itself, also depends on surface humidity and micro-roughness \cite{Franceschi09,Nield11,lidar_surface_reflectance}).
In simple cases for which the distance and the incidence angle are
not greatly changing (cliff survey for instance), the laser intensity
can be used to distinguish between materials relatively efficiently~\cite{Franceschi09}.
It can also improve the robustness of a classifier based on simple
geometrical parameters (\cite{Guarnieri09}) in the case of simple
natural environments such as riparian vegetation over sand. However,
for complex scenes such as the Otira river, lidar intensity is more
difficult to use given the large changes in distances, incidence angles
and state of the surface (wet vs dry surfaces). Moreover, no standard
exists between scanner manufacturers such that even if surface reflectance
could be isolated from other effects, it would not necessarily be
comparable between various scanner measurements as opposed to purely
geometrical descriptors that can be used for any source of data (i.e.
classifiers can be exchanged between users independently of the scanner
used to acquire the data). Because laser reflected intensity is not
globally nor temporally consistent on a complex 3D scene, we conclude
that it cannot yet be used as a primary classifier of complex 3D natural
scenes. Development of precise geometrical corrections factors for
reflected intensity (e.g., \cite{Correction_methods_intensity_2011})
may allow its future inclusion in the process of classifier training
and subsequent classification to improve the resolution and accuracy
of the geometrical classification. Provision for this is already included
in the software.

In the case of airborne lidar, the combination of geometrical information
and imagery can significantly increase classification quality (e.g.
\cite{fusion_hypersectral_lidar}). However, RGB imagery have rarely
been used in the context of 3D terrestrial lidar classification \cite{Lichti05}.
The main reason is that it is much more difficult to have a spatially
consistent RGB imagery of a 3D complex scene from the ground than
from air. Indeed, the more 3D and complex is an outdoor scene, the
more difficult it is to get a consistent exposure from the ground
and from different points of view. For instance, in the case of the
Otira River, the extent of shadows is pronounced owing to the narrow
gorge configuration and to the presence of vegetation, and variable
during one day. Wetted surfaces which are common in fluvial environments
also have a different spectral signature than dry surfaces. Also,
RGB imagery cannot distinguish first or last laser reflexions in the
case of the new generation of full-waveform multi-echo scanners. However,
in the context of the riparian vegetation case example (fig. \ref{fig:bench_visu}),
and owing to the strong difference in spectral characteristics of
the vegetation and the sandy bed, good success could be expected using
RGB classification \cite{Lichti05}. But this requires the imagery
to be taken without strong shadows, and in the case of the Leica Scanstation
2 would be limited by the low resolution of the on-board camera and
the lack of precise registration with the point cloud (typically a
few cm difference at 50 m). Classifying flat versus ripples zones
would still require a geometrical analysis.

Because it works in 3D, our method can be used on 2.5D airborne lidar
or mostly 2D point clouds (\cite{Guarnieri09,HodgeESPL09,Milan10,BareEarth}).
As shown with the riparian vegetation example, it allows a direct
extraction of vegetation on the raw data without the need to construct
a raster DEM. Because of the smaller density of points and the smaller
range of scales available to characterize trees in full waveform aerial
lidar than in ground based lidar, it is not certain that the method
will perform significantly better in defining the ground than existing
ones for aerial data. However, it should perform well as a generic
geometric classifier of surfaces. Although not used in this study,
the multiscale calculation also records the vertical angle of the
local surface at the largest scale. This could be used as an additional
constraint to detect buildings from road for instance.

\section{Conclusion}

We introduced a new method for classifying 3D point clouds, called
multi-scale dimensionality analysis, to characterize features according
to their geometry. We demonstrated the applicability of this method
in two contexts: 1. separating riparian vegetation from the ground
in the Mt St Michel bay, and 2. recognizing rocks, vegetation, water
and gravels in a steep mountain river bed. In each case the classification
was performed with very good accuracy. The method is robust to missing
data and changes in resolution common in ground-based lidar data.
By combining various scales, the method systematically performs better
than a single scale analysis and improves the spatial resolution of
the classification.

Multi-scale dimensionality analysis proves quite efficient especially
in separating vegetation from the rest of the data. Removing (or studying)
vegetation is a common issue in natural environment studies and this
method will be useful in this context given that it can work directly
on the raw data. Typical application include bare ground detection
to study sedimentation/erosion patterns in fluvial environments (\cite{Wheaton09}),
rock face analysis on which vegetation can grow and create unwanted
noise (\cite{Labourdette07,RosserJGR07}) or analysis of vegetation
patterns and their relation to hydro-sedimentary processes.

We gave a particular attention to provide tools usable by non-specialists
of machine learning, while retaining the ability to process large
batches of data automatically. This tool set is available as Free/Libre
software on the \href{http://nicolas.brodu.numerimoire.net/}{first author home page}%
\footnote{See \href{http://nicolas.brodu.numerimoire.net/en/recherche/canupo/}{http://nicolas.brodu.numerimoire.net/en/recherche/canupo/}
(as of 2012/01/19)%
}. Because it relies only on geometrical properties, classifier parameter
files can be exchanged between users and applied on any geometrical
data without going over the process of training the classifier.

\section{Acknowledgments}

Daniel Girardeau-Montaut is greatly acknowledged for his ongoing development
of the free point-cloud analysis and visualization software CloudCompare
\cite{CloudCompare} used in this work for visualization and point
cloud segmentation according to their class. These activities have
been carried out with the support of the European Research Executive
Agency in the framework of the Marie Curie International Outgoing
Fellowship €ROSNZ for D. Lague, and with the support of the \href{http://risc-e.univ-rennes1.fr}{RISC-E}
(Interdisciplinary Research network on Complex Systems in Environment)
project for N. Brodu. 

\bibliographystyle{plain}
\bibliography{canupo}

\end{document}